\documentclass[conference]{IEEEtran}
\IEEEoverridecommandlockouts
\usepackage{cite}
\usepackage{amsmath,amssymb,amsfonts}
\usepackage{algorithmic}
\usepackage{graphicx}
\usepackage{textcomp}
\usepackage{float}                  
\usepackage{subfig}
\usepackage{overpic}
\usepackage{xcolor}
\usepackage{epstopdf}
\usepackage{booktabs}
\usepackage{color}
\usepackage{balance}
\usepackage{array}
\usepackage{textcomp}
\usepackage{hyperref}

\def\BibTeX{{\rm B\kern-.05em{\sc i\kern-.025em b}\kern-.08em
    T\kern-.1667em\lower.7ex\hbox{E}\kern-.125emX}}
\begin{document}


\title{%
  {\normalfont\footnotesize\mdseries
   \parbox{\textwidth}{\centering
     This work has been submitted to the IEEE for possible publication. Copyright may be transferred without notice, after which this version may no longer be accessible.
   }}
  \vspace{0.1cm}\\
  Low-Cost Mobile Bimanual Manipulation for Long-Horizon Toilet Cleaning
}

\author{
Sven Schreiber$^{1\dagger}$,
Bernard Burdiek$^{1\dagger}$,
Lasse Huber-Saffer$^{1\dagger}$,
Yung-Hui Li$^{2}$,
Donglin Chen$^{3*}$,\\
Shang-Ching Liu$^{1*}$,
Jianwei Zhang$^{1}$
\thanks{
$^{1}$Technical Aspects of Multimodal Systems (TAMS), Department of
Informatics, University of Hamburg;
$^{2}$Hon Hai Research Institute (HHRI);
$^{3}$National Frontiers Science Center for Industrial Intellignence and Systems Optimization, Northeastern University, China and the Key Laborator of Data Analytics and Optimization for Smart Industry (Northeastern University), Ministry of Education, China
$^{\dagger}$These authors contributed equally.
$^{*}$Corresponding authors: Donglin Chen
(2210284@stu.neu.edu.cn) and Shang-Ching Liu
(shang-ching.liu@uni-hamburg.de).
}
}

\maketitle
\begin{abstract}
Vision-Language-Action (VLA) models have shown promising capabilities
for robot manipulation, but their performance on accessible mobile
bimanual platforms remains underexplored. We present a low-cost and
reproducible mobile bimanual platform integrating two LeRobot SO-101
manipulators with a TurtleBot~4 and a LeRobot--ROS~2 software stack.
Using long-horizon toilet cleaning as a case study, we collect 250
real-world demonstrations and evaluate $\pi_0$, $\pi_{0.5}$, SmolVLA,
and GR00T under the same embodiment and task conditions. We further
evaluate variants using SARM-derived task-progress signals and
short-term MEM observation history. Among the evaluated models,
$\pi_0$ performs best, achieving 60\% strict and 70\% relaxed
autonomous success, increasing to 90\% with limited human intervention.
In contrast, the evaluated progress- and memory-augmented variants do
not improve their corresponding baselines. Our results demonstrate the
feasibility of long-horizon mobile bimanual manipulation on accessible
hardware while highlighting remaining challenges in reliable VLA
deployment. Project website: \url{https://so101-turtlebot4.github.io/mobile-bimanual-cleaning/}.
\end{abstract}


\section{Introduction}

\begin{figure}[!t]
    \centering
    \includegraphics[height=6cm]{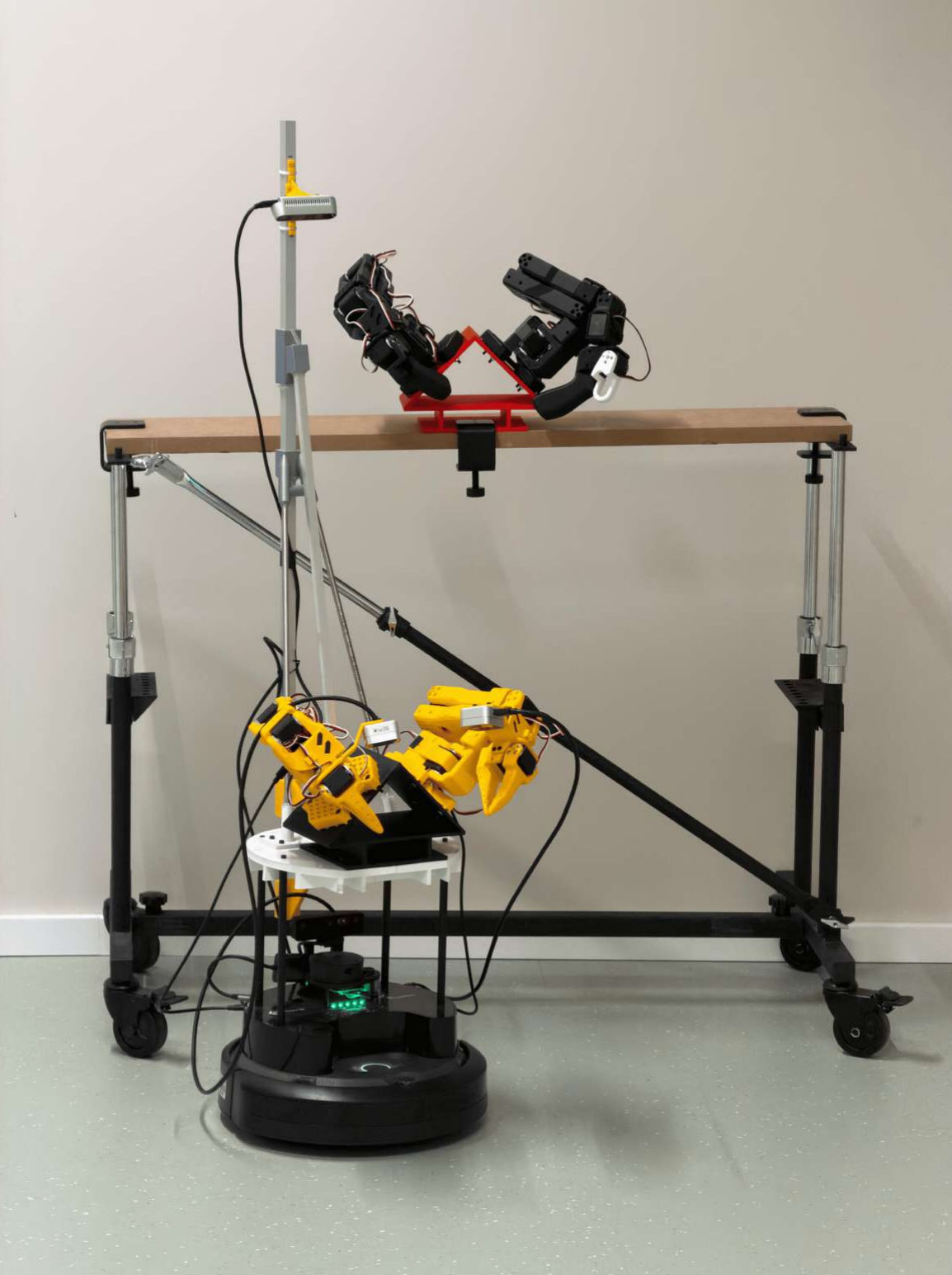}
    \hspace{0.3cm}
    \includegraphics[height=6cm]{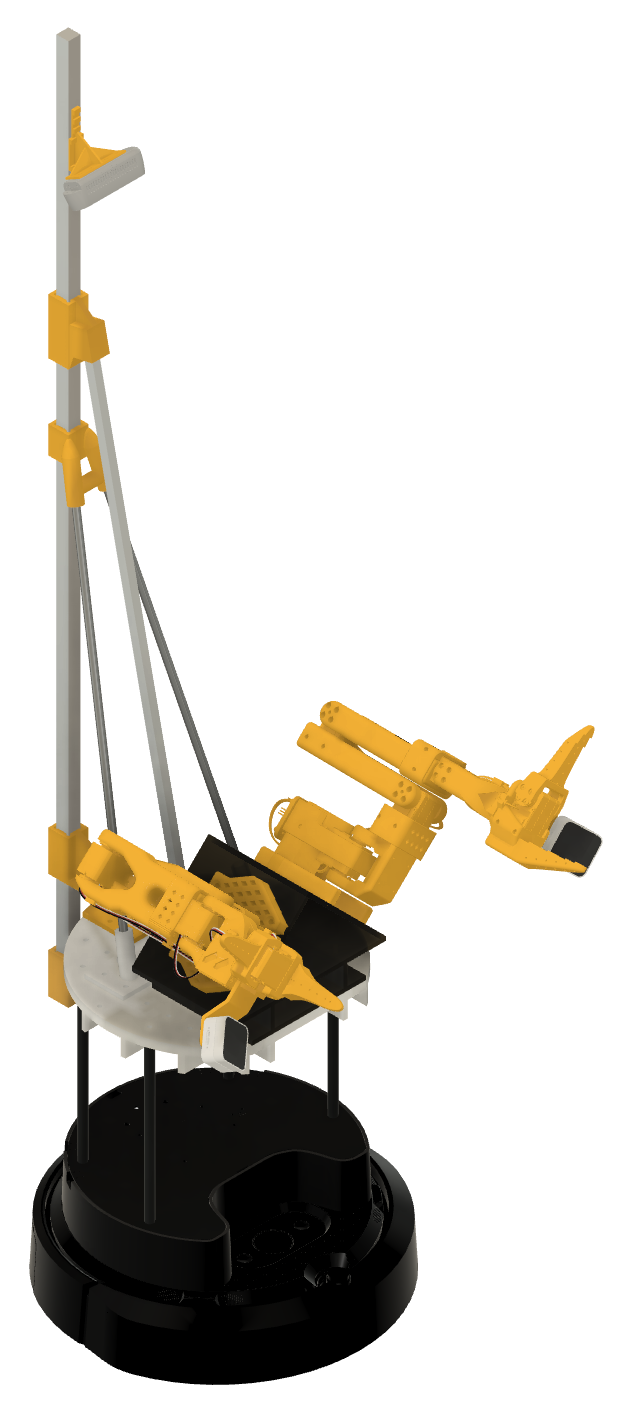}
    \caption{The low-cost mobile bimanual robot.}
    \label{FigRobot}
\end{figure}

Vision-Language-Action (VLA) models have rapidly advanced from
task-specific manipulation toward generalist robot policies capable of
dexterous, bimanual, and increasingly long-horizon behaviors~\cite{zitkovich2023rt2,kim2024openvla,black2025pi0,black2025pi05}.
However, translating these advances into reproducible real-world
systems remains challenging~\cite{li2024evaluating}. Existing models are typically evaluated
on different robot embodiments, datasets, and task settings, making it
difficult to understand how they compare when deployed under the same
physical conditions~\cite{li2024evaluating,guruprasad2024benchmarking}. This challenge is particularly pronounced for
mobile bimanual manipulation, where locomotion, dual-arm coordination,
perception, and contact-rich manipulation must operate reliably over
extended task sequences~\cite{fu2024mobile}.
In this work, we present a low-cost mobile bimanual platform for
studying VLA policies in such settings, as shown in Fig. \ref{FigRobot}. The platform combines two
LeRobot SO-101 manipulators with a TurtleBot~4 mobile base, custom
3D-printable mechanical structures, and a multi-camera perception
system. A modular LeRobot--ROS~2 software stack supports whole-robot
teleoperation, data collection, human-in-the-loop correction, policy
training, and deployment. The system is designed to provide an
accessible and reproducible testbed for physical evaluation of modern
VLA policies.
We evaluate the platform through a long-horizon toilet-cleaning case
study that requires mobile navigation, bimanual coordination, tool
manipulation, contact-rich cleaning, and sequential task execution.
Using 250 real-world demonstrations collected on the same platform, we
fine-tune and physically evaluate representative open VLA models,
including $\pi_0$~\cite{black2025pi0}, $\pi_{0.5}$~\cite{black2025pi05}, SmolVLA~\cite{shukor2025smolvla}, and GR00T N1.7~\cite{bjorck2025gr00tn1,nvidia2026gr00tn17}. We additionally
evaluate two temporal variants based on SARM-derived task-progress
signals~\cite{chen2026sarm} and short-horizon MEM observation history~\cite{torne2026mem}.

Our experiments reveal substantial differences between VLA
configurations under identical physical conditions. Among the evaluated
models, $\pi_0$ achieves the strongest autonomous performance, whereas
the evaluated SARM- and memory-augmented variants do not improve task
success in our setting. The experiments further expose recurring
failure modes in embodiment adaptation, stage transitions, and
long-horizon execution, while showing that limited human intervention
can recover a substantial fraction of otherwise unsuccessful
$\pi_0$ rollouts. These results highlight the remaining gap between
the capabilities demonstrated by modern VLA models and reliable
deployment on accessible mobile bimanual hardware.

The main contributions of this work are:
\begin{itemize}
    \item We present a low-cost and reproducible mobile bimanual
    platform based on dual LeRobot SO-101 manipulators and TurtleBot~4,
    with custom 3D-printable structures, multi-camera perception, and
    an integrated LeRobot--ROS~2 software stack.

    \item We establish a real-world long-horizon toilet-cleaning case
    study with 250 demonstrations and five sequential task stages,
    requiring mobility, bimanual coordination, and contact-rich tool
    manipulation.

    \item We provide a controlled physical evaluation of representative
    open VLA models and temporal variants on the same embodiment and
    dataset, and analyze their task performance and characteristic
    failure modes.
\end{itemize}

\section{Related Work}

\subsection{Vision-Language-Action Models for Generalist Manipulation}

Large-scale robot learning has increasingly shifted from task-specific
policies toward generalist models that jointly leverage visual
observations, language instructions, and robot actions. RT-1
demonstrated that Transformer-based policies can acquire a broad
repertoire of real-world manipulation skills~\cite{brohan2022rt1}, while RT-2 further
showed that semantic knowledge acquired through vision-language
pretraining can be transferred to robotic control~\cite{zitkovich2023rt2}. Open
X-Embodiment subsequently introduced large-scale cross-embodiment robot
datasets and policies~\cite{o2024open}, motivating open generalist models such as
Octo~\cite{ghosh2024octo} and OpenVLA~\cite{kim2024openvla}. Together, these works established the
feasibility of training reusable robot policies across diverse tasks,
environments, and embodiments.

More recent VLA models have increasingly targeted dexterous, bimanual,
and long-horizon manipulation. RDT-1B introduced a diffusion foundation
model specifically designed for bimanual manipulation~\cite{liu2025rdt}. The
$\pi_0$ model employs flow matching for continuous action generation
and demonstrates diverse dexterous manipulation behaviors across
multiple robot embodiments~\cite{black2025pi0}. Its successor, $\pi_{0.5}$, further
targets open-world generalization through heterogeneous co-training
and high-level semantic prediction, demonstrating multi-stage mobile
manipulation in previously unseen household environments~\cite{black2025pi05}.
SmolVLA explores a complementary direction toward compact and
computationally efficient VLA models, with a particular emphasis on
affordable robotic hardware and community-collected datasets~\cite{shukor2025smolvla}.
The GR00T family instead develops generalist foundation models for
humanoid and bimanual robotic embodiments~\cite{bjorck2025gr00tn1}.

Although these models report strong results on their respective
platforms and benchmarks, their evaluations differ substantially in
robot embodiment, training data, task definition, and experimental
protocol. Consequently, reported performance across individual studies
does not directly characterize how different VLA families behave when
adapted to the same low-cost mobile bimanual embodiment and trained on
the same real-world demonstrations. In this work, we use a common
physical platform, dataset, and long-horizon task to provide a
controlled case study of several representative VLA families,
including $\pi_0$, $\pi_{0.5}$, SmolVLA, and GR00T.

\subsection{Low-Cost Bimanual and Mobile Manipulation}

In parallel with advances in learning algorithms, accessible robotic
hardware has substantially lowered the barrier to collecting
large-scale manipulation data and evaluating learned policies. ALOHA
introduced a low-cost and open-source bimanual teleoperation system
for fine-grained manipulation \cite{zhao2023learning}, while ALOHA 2 subsequently improved
the robustness, ergonomics, and maintainability of the platform \cite{aldaco2024aloha}.
ALOHA Unleashed further demonstrated that large-scale imitation
learning on accessible bimanual hardware can produce dexterous
behaviors involving deformable objects and complex contact
interactions \cite{zhao2024aloha}.

Mobile ALOHA extended this paradigm from tabletop manipulation to
whole-body mobile bimanual operation by combining an ALOHA-based
bimanual system with a mobile base and whole-body teleoperation
interface \cite{fu2024mobile}. The platform demonstrated autonomous household
behaviors involving coordinated locomotion and manipulation, including
cooking, cabinet interaction, elevator operation, and rinsing
cookware. These results established that relatively accessible
hardware can support complex mobile bimanual learning and provided an
important foundation for studying whole-body policies outside
fixed-base workspaces.

The LeRobot ecosystem has recently provided another route toward
accessible robot learning by integrating standardized datasets,
teleoperation, policy training, and deployment within an open-source
framework \cite{cadene2026lerobot}. Low-cost serial manipulators such as the SO-100 and
SO-101 further reduce the hardware requirements for collecting
real-world demonstrations and evaluating learned policies \cite{Knight_Standard_Open_SO-100}.
However, most existing LeRobot-based platforms and datasets focus on
fixed-base manipulation, while mobile bimanual configurations remain
less extensively characterized as physical testbeds for generalist VLA
policies.

Our platform follows the broader direction established by ALOHA and
Mobile ALOHA, but targets a complementary design point centered on the
LeRobot ecosystem. We integrate two SO-101 manipulators with a
TurtleBot~4 mobile base and a multi-camera perception system, together
with custom 3D-printable mechanical structures and a modular
LeRobot--ROS~2 software stack. The resulting system is intended as an accessible and reproducible
experimental platform for data collection, human-in-the-loop
correction, and physical evaluation of VLA policies under mobile and
bimanual task requirements.

\subsection{Long-Horizon Manipulation, Progress Estimation, and Memory}

Long-horizon manipulation introduces additional challenges beyond
generating locally appropriate actions. During multi-stage execution,
a robot must maintain consistency across successive behaviors,
transition between subtasks, and account for information that may no
longer be evident from the current observation. Recent work has
therefore investigated mechanisms for representing task progress and
maintaining temporal context during robot execution.

One direction explicitly models progress through a task. Stage-Aware
Reward Modeling (SARM) learns a video-based reward model that jointly
estimates high-level task stages and fine-grained progress within each
stage \cite{chen2026sarm}. By deriving supervision from natural-language subtask
annotations, SARM provides a structured representation of task
progress that can be used to evaluate demonstrations and improve
policy learning through Reward-Aligned Behavior Cloning.
Long-VLA approaches have similarly investigated phase-aware
representations and subtask structure to improve execution over
extended manipulation sequences \cite{fan2025long}.

A complementary direction incorporates observation or interaction
history directly into the policy. MemoryVLA introduces explicit memory
mechanisms for retrieving task-relevant historical information during
action generation \cite{shi2026memoryvla}, while OptimusVLA employs global and local memory
to improve temporal consistency and infer execution progress \cite{li2026global}.
Multi-scale Embodied Memory (MEM) further combines short-term visual
history with abstract long-term memory for extended VLA execution \cite{torne2026mem}.
Such mechanisms have demonstrated the potential of temporal context for
tasks whose current state cannot be completely inferred from a single
observation.

Rather than proposing a new progress or memory mechanism, we use these
directions as part of our empirical study. In addition to the standard
VLA configurations, we evaluate two temporal variants available within
or compatible with the LeRobot ecosystem. The first augments
$\pi_{0.5}$ with short-term observation memory based on MEM, testing
whether recent visual history improves long-horizon execution. The
second augments $\pi_0$ observations with task-stage and progress
predictions produced by SARM, testing whether explicit progress
information facilitates transitions between subtasks. Using these variants 
we examine whether straightforward access to additional temporal information 
translates into improved physical task performance.

Taken together, prior work has independently advanced generalist VLA
policies, accessible mobile bimanual hardware, and mechanisms for
long-horizon temporal awareness. Our work provides a physical case
study at the intersection of these directions. Using a low-cost
LeRobot--TurtleBot~4 platform and a common real-world dataset, we
compare representative open VLA implementations on a multi-stage
toilet-cleaning task and analyze their performance and failure modes,
including whether memory- and progress-augmented variants improve
long-horizon execution in this setting.
\section{Mechanical Design}
To align with the primary objective of developing a low-cost mobile manipulation platform while maximizing reproducibility for the robotics community, we adopt two LeRobot SO-101 robotic arms \cite{cadene2026lerobot} for bimanual manipulation and a TurtleBot~4 as the mobile base. 
Specifically, several structural components are redesigned to improve mechanical integration, payload support, and sensor placement. 
By relying exclusively on readily accessible hardware and additively manufactured custom parts, researchers and practitioners can reproduce the entire robotic platform with minimal effort and expense.
\begin{figure}[htbp]
	\centering
	\includegraphics[width=.5\columnwidth]{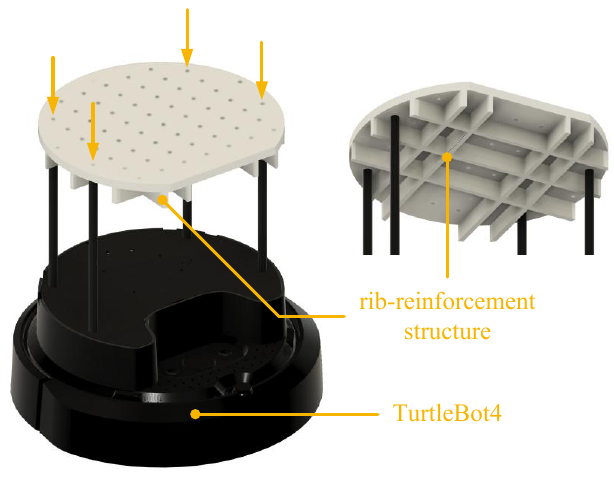}
	\caption{Redesigned top cover. The four mounting holes indicated by the yellow arrows preserve the original TurtleBot~4 mounting interfaces.}
    \label{FigTopCover}
\end{figure}
\par We redesigned a new top cover for TurtleBot~4, as shown in Fig. \ref{FigTopCover}, to enhance its expandability and payload capacity. 
Serving as the foundation of the upper mechanical assembly, the redesigned top cover preserves the original mounting interfaces to ensure seamless integration with the TurtleBot~4 while introducing numerous additional mounting holes for attaching the customized components.
\begin{figure}[htbp]
	\centering
	\includegraphics[width=.5\columnwidth]{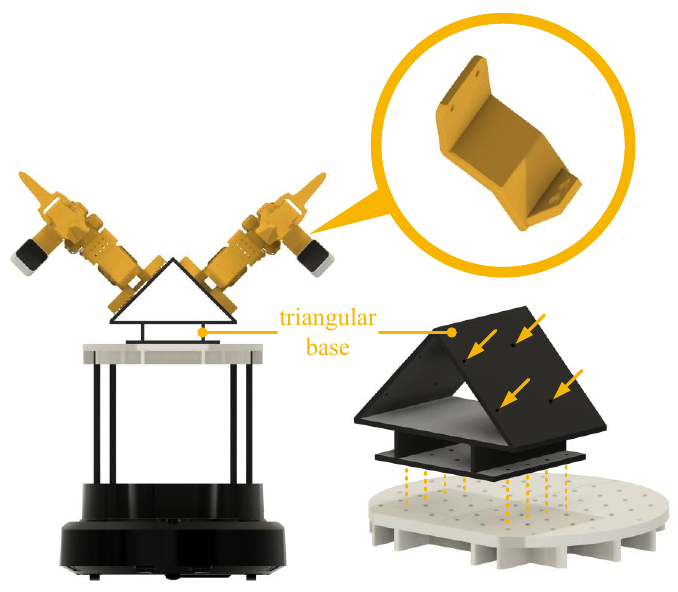}
	\caption{Triangular base. The enlarged inset shows the holder for an Intel RealSense D405 camera. The yellow arrows indicate the mounting holes aligned with the base of a SO-101 arm.}
    \label{FigTriangularBase}
\end{figure}
\par As shown in Fig. \ref{FigTriangularBase}, a customized triangular base was mounted above the redesigned top cover to support the two SO-101 arms while reducing workspace interference between the arms. 
\par
\begin{figure}[htbp]
	\centering
	\includegraphics[width=0.5\columnwidth]{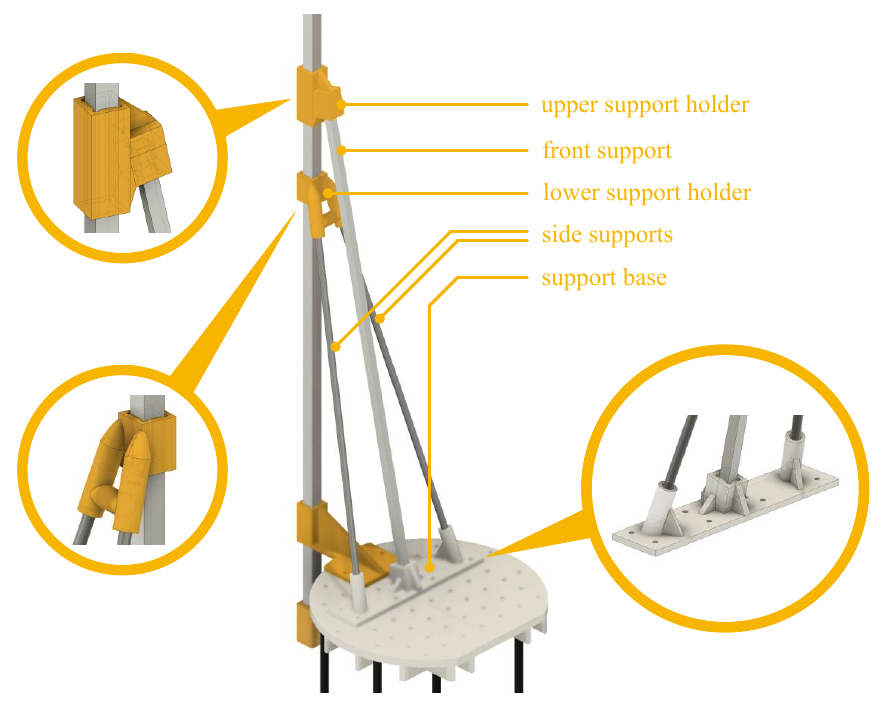}
	\caption{The support structures comprise an upper support holder, a lower support holder, a front support, two side supports, and a support base.}
    \label{FigSupport}
\end{figure}

 A commercially available $15~\mathrm{mm} \times 15~\mathrm{mm} \times 1~\mathrm{m}$ hollow aluminum tube is adopted as the support for the top camera due to its low cost, widespread availability, and lightweight characteristics. 
Two additional support structures were introduced between the aluminum tube and the redesigned top cover, as illustrated in Fig. \ref{FigSupport}. 
These supports substantially improve the stiffness of the camera mounting assembly, thereby suppressing camera oscillations during robot operation.
\par 
The complete platform costs approximately \texteuro{}3200. The TurtleBot~4 accounts for approximately \texteuro{}2700, while the complete mobile bimanual manipulation extension costs approximately \texteuro{}500, including two SO-101 follower arms, two leader arms for teleoperation, the RGB camera setup, mechanical hardware, and the custom 3D-printed components.
To quantitatively demonstrate the affordability and reproducibility of the proposed robot, Table \ref{TabCost} summarizes the manufacturing statistics of all customized 3D-printed structural components, including the printing time, filament consumption, and estimated printing cost. 
In total, the customized parts require approximately 71 h 53 min of sequential printing time and consume 268.73 m of filament, with a total estimated manufacturing cost of \texteuro{}22.5. 
\begin{table}[htbp]
     \centering
    \caption{Statistics of the customized 3D-printed components.}
    \label{TabCost}
    \scriptsize
    \setlength{\tabcolsep}{4pt}
    \renewcommand{\arraystretch}{1.1}
    \begin{tabular}{
        >{\centering\arraybackslash}m{0.14\columnwidth}|
        >{\centering\arraybackslash}m{0.15\columnwidth}
		>{\centering\arraybackslash}m{0.1\columnwidth}|
		>{\centering\arraybackslash}m{0.14\columnwidth}|
		>{\centering\arraybackslash}m{0.15\columnwidth}
		>{\centering\arraybackslash}m{0.1\columnwidth}
    }
        \toprule
        Component & Time & Length & Component & Time & Length \\
		\midrule
        \includegraphics[height=0.8cm]{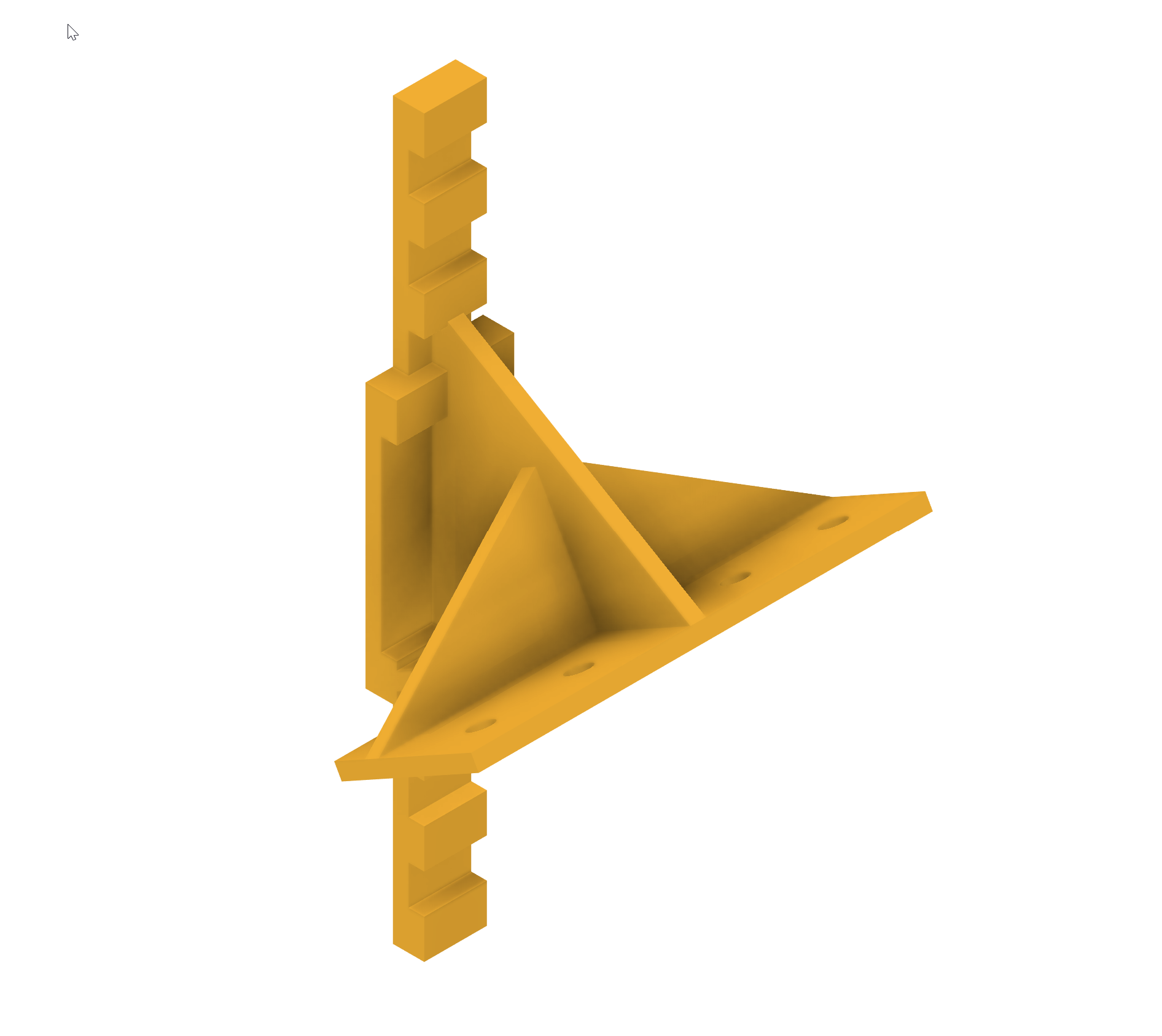} & 1 h 33 min & 4.4 m &
		\includegraphics[height=0.8cm]{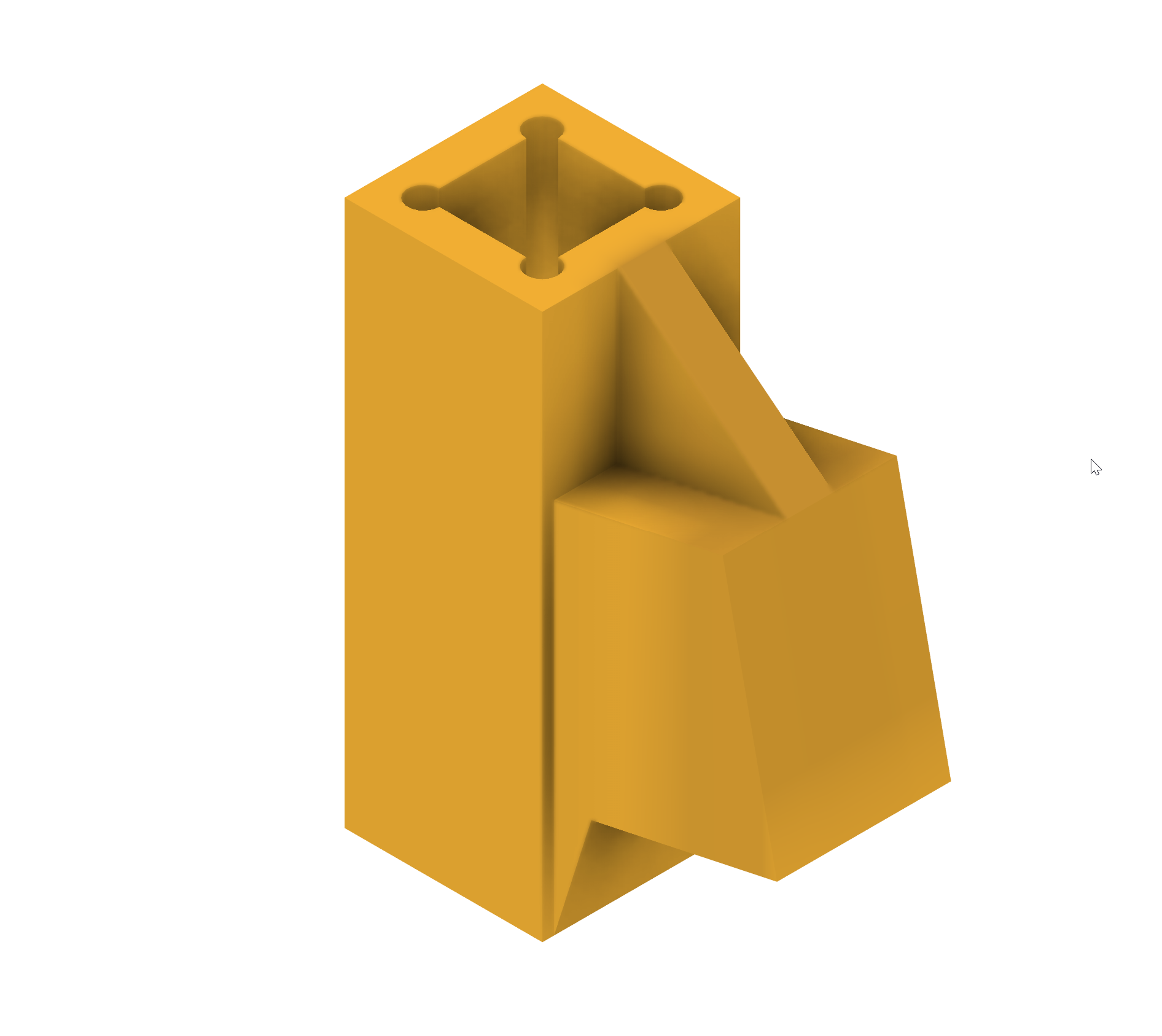} & 2 h 47 min & 10.74 m\\
        \midrule
		\includegraphics[height=0.8cm]{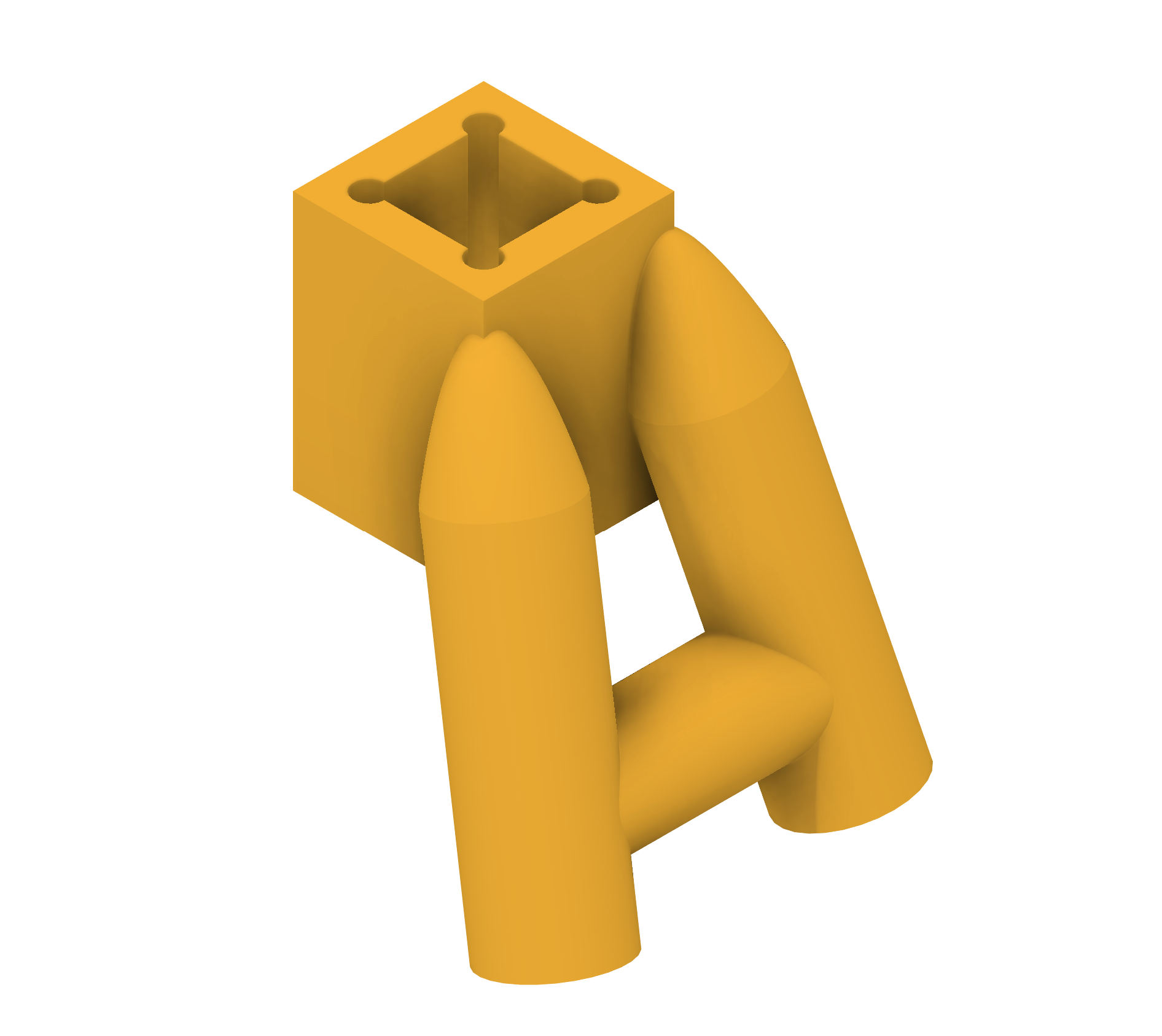} & 3 h 54 min & 12.07 m &
		\includegraphics[height=0.8cm]{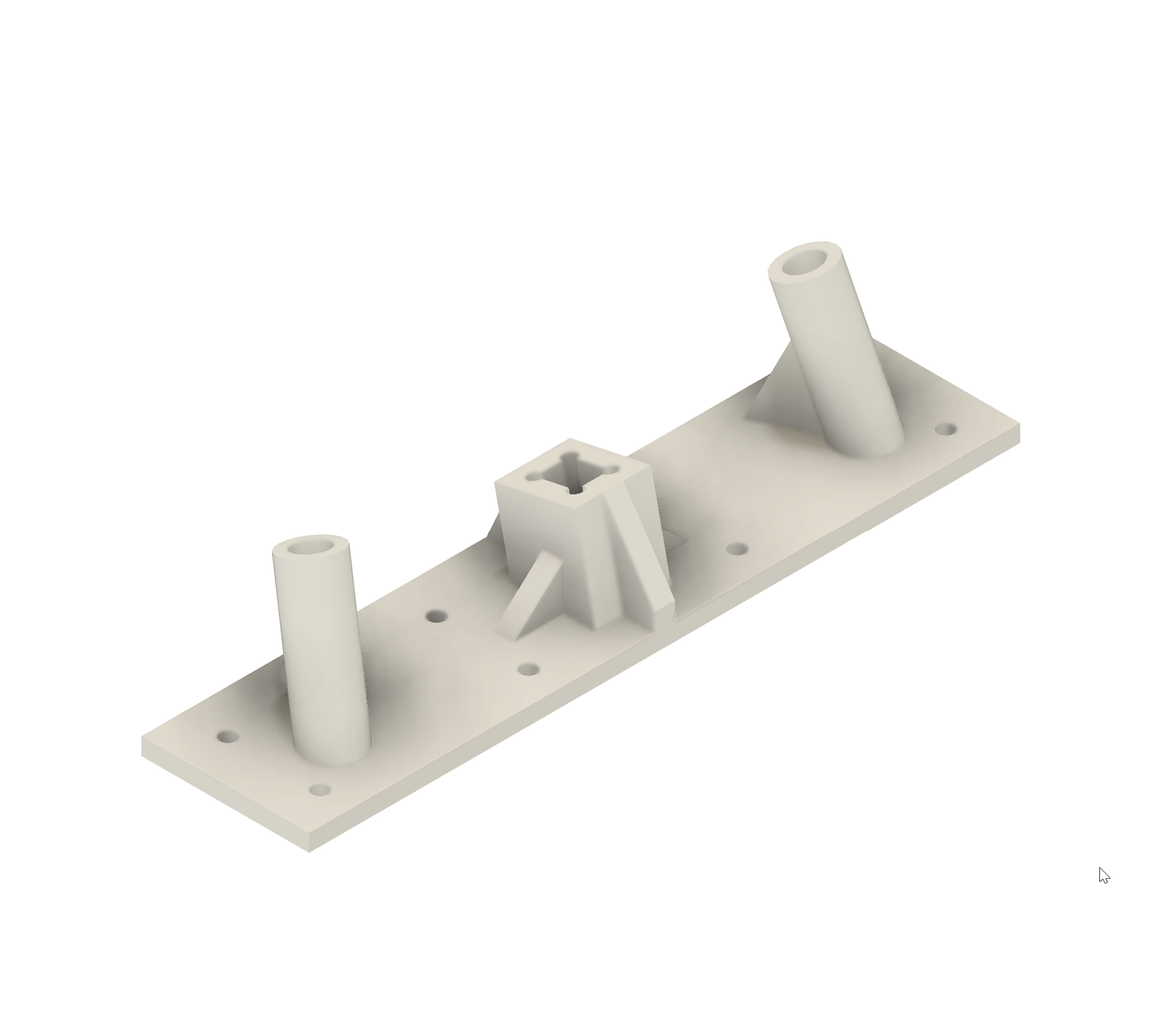} & 4 h 20 min & 16.00 m\\
        \midrule
		\includegraphics[height=0.8cm]{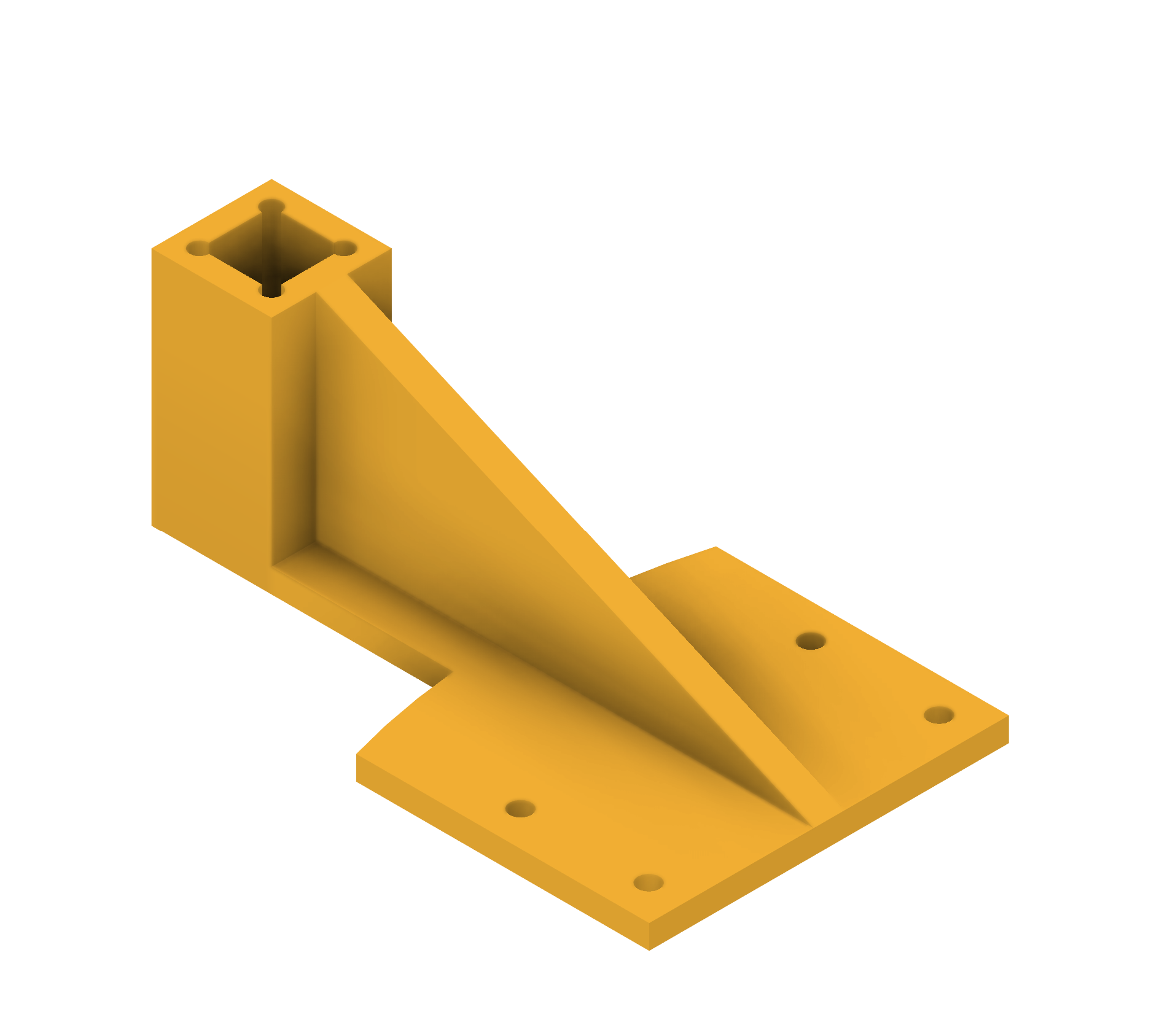} & 3 h 24 min & 12.98 m &
		\includegraphics[height=0.8cm]{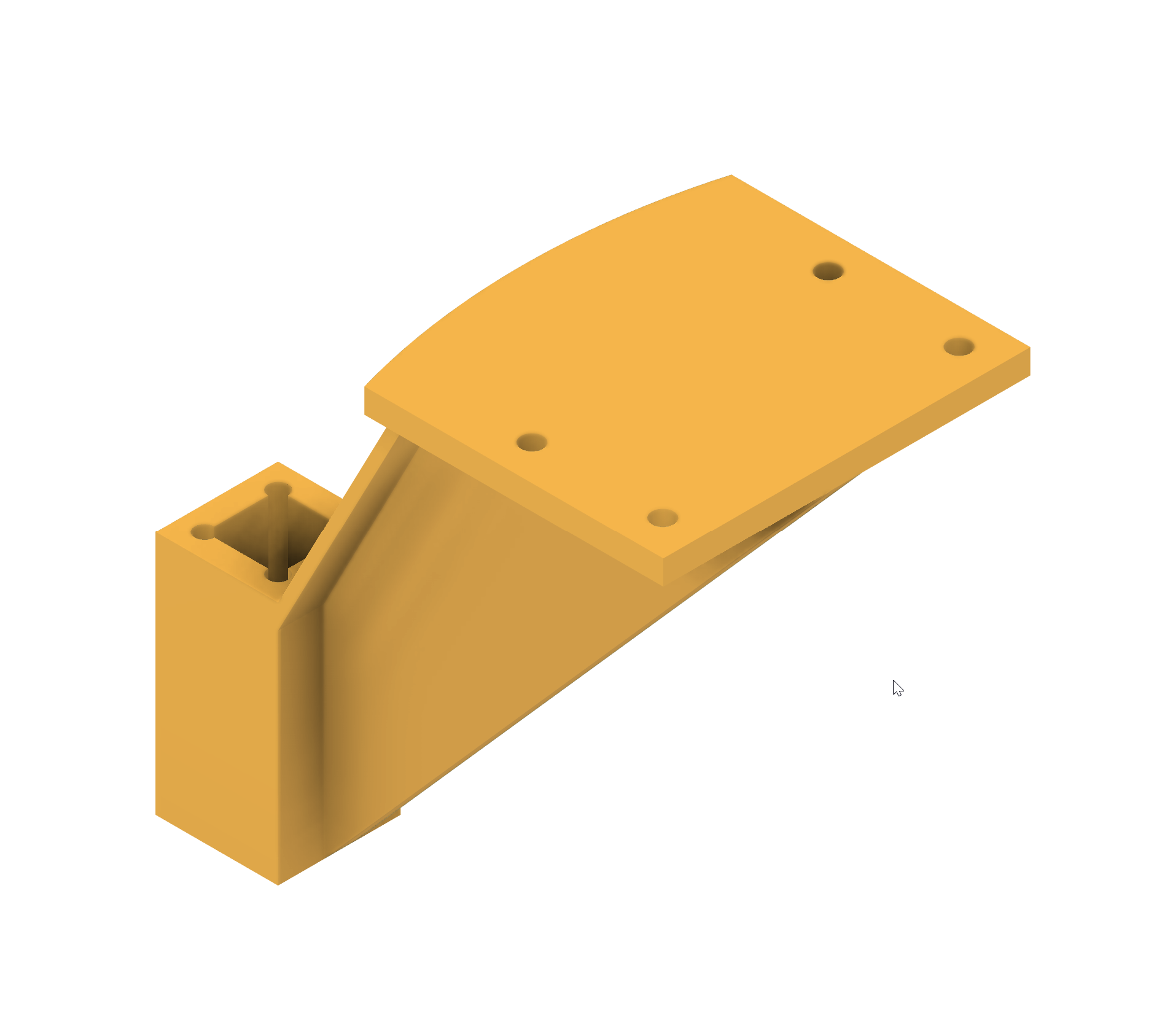} & 5 h 42 min & 21.13 m\\
		\midrule
		\includegraphics[height=0.8cm]{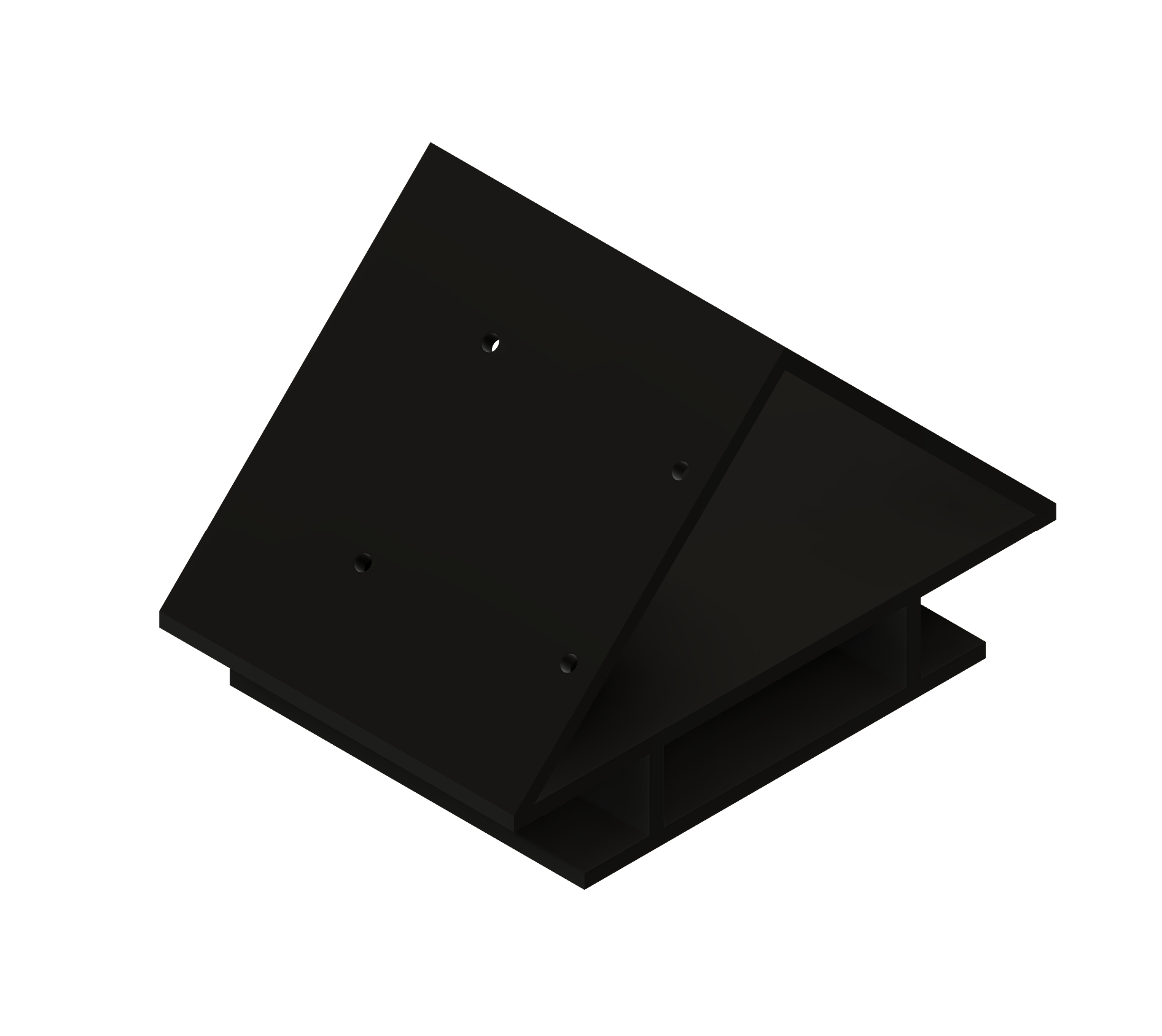} & 24 h 28 min & 83.06 m &
		\includegraphics[height=0.8cm]{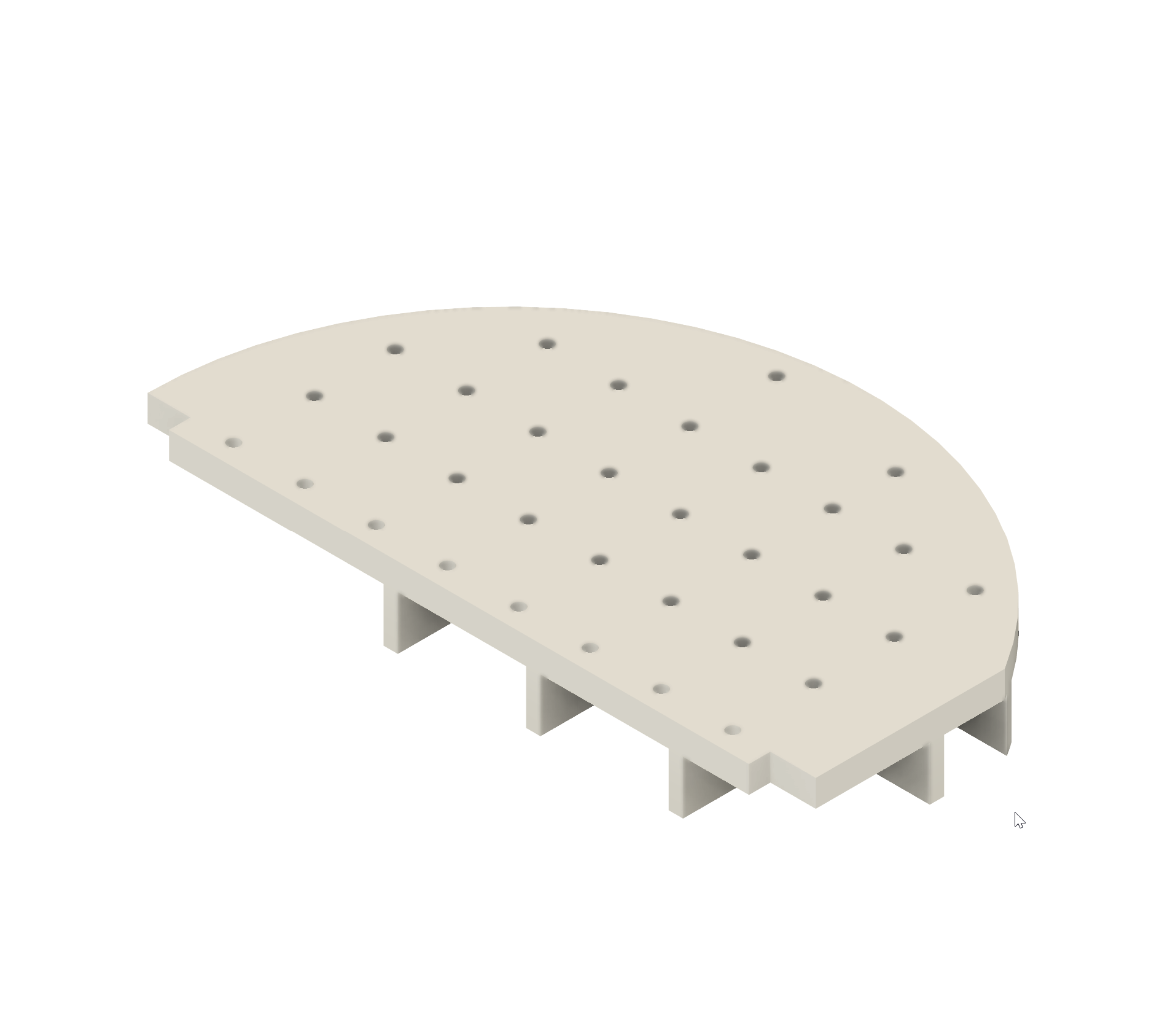} & 13 h 33 min & 57.07 m\\
        \midrule
		\includegraphics[height=0.8cm]{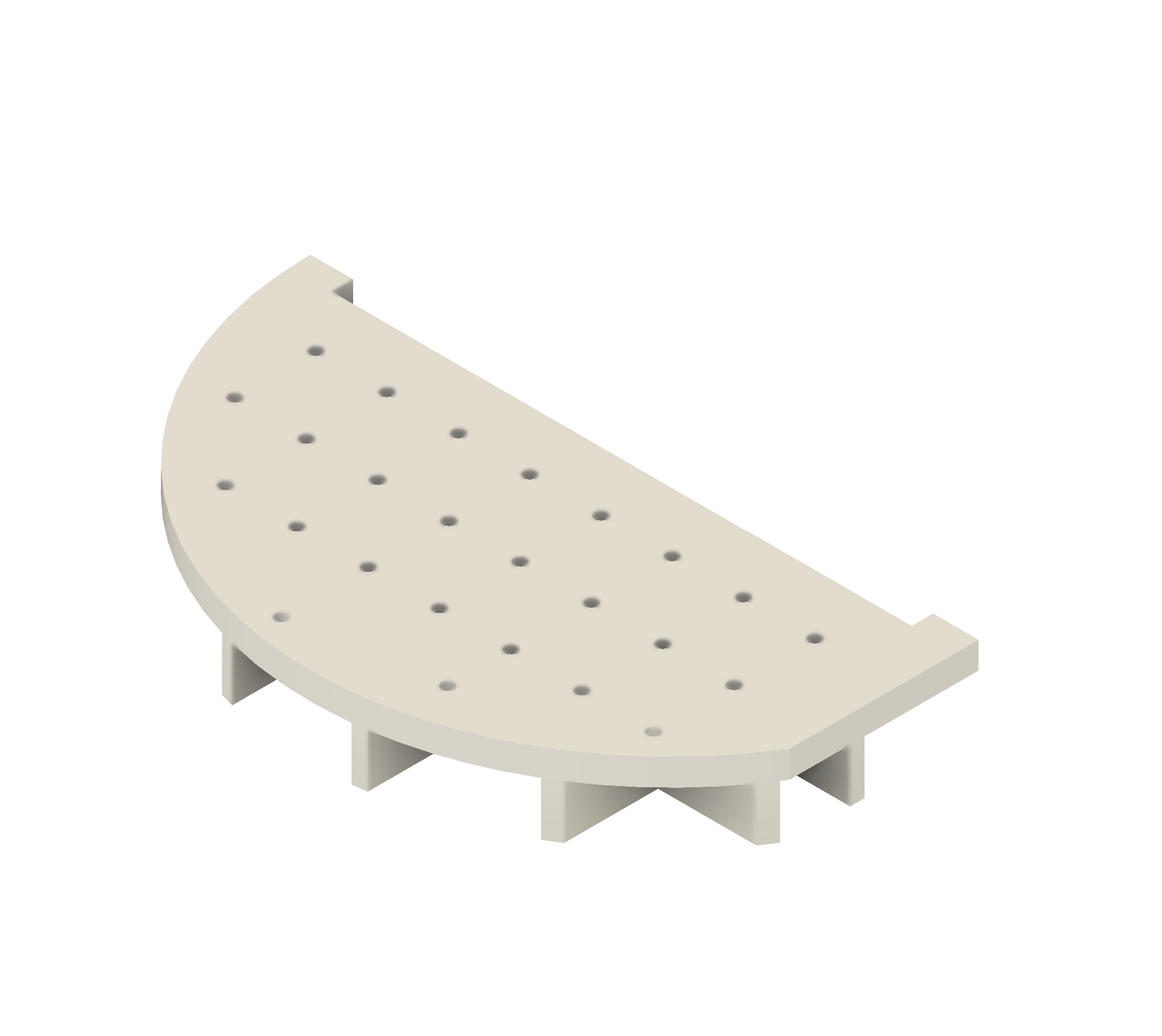} & 12 h 12 min & 51.28 m &
        - & - & -\\
        \bottomrule
    \end{tabular}
\end{table}

\section{Software Design}
The robot and teleoperation stack is tightly integrated with the LeRobot~\cite{cadene2026lerobot} ecosystem, enabling rapid iteration by reusing existing infrastructure for teleoperation, data collection, policy training, and policy rollouts. Communication between the robot and host workstation is handled through a Zenoh–ROS 2 bridge, providing efficient network transport and supporting operation of multiple robots on the same network.

On the host, a LeRobot adapter subscribes to ROS 2 topics for camera observations, odometry, and joint states, and publishes whole-robot action commands. A separate onboard ROS 2 node interfaces directly with the two SO-101 arms. This separation allows computationally intensive policy inference to run on a workstation, while the onboard Raspberry Pi 4 remains responsible for real-time hardware communication and low-level control.

The bimanual platform exposes an action and observation space consisting of five joints per arm plus one actuated gripper joint, linear and angular mobile-base velocities, and three RGB camera views: a top-mounted camera and one wrist-mounted camera on each arm.

Our teleoperation setup uses two SO-101 leader arms for arm control and a keyboard for controlling the mobile base. Although we considered foot pedals, the episode recording scripts required additional controls, so we opted for two-person teleoperation: one operator controls the arms, while the other controls the base and episode recording. The same interface supports torque control, which is particularly useful for interactive data-collection methods such as DAgger~\cite{ross2011reduction}, where control may be handed between the learned policy and a human operator.

This modular design makes the platform well suited for both rapid prototyping and educational use. Individual components, such as cameras, manipulators, teleoperation interfaces, or learned policies, can be modified or replaced without redesigning the complete system. Building on LeRobot and ROS 2 also gives users access to familiar abstractions and existing tooling, which lowers the barrier to experimenting with data collection, imitation learning, and mobile manipulation. As a result, the same platform can support quick research iterations as well as hands-on robotics and machine-learning projects in teaching environments.

\section{Experiments} 
\begin{figure*}[htbp]
    \centering
	\includegraphics[width=0.65\textwidth]{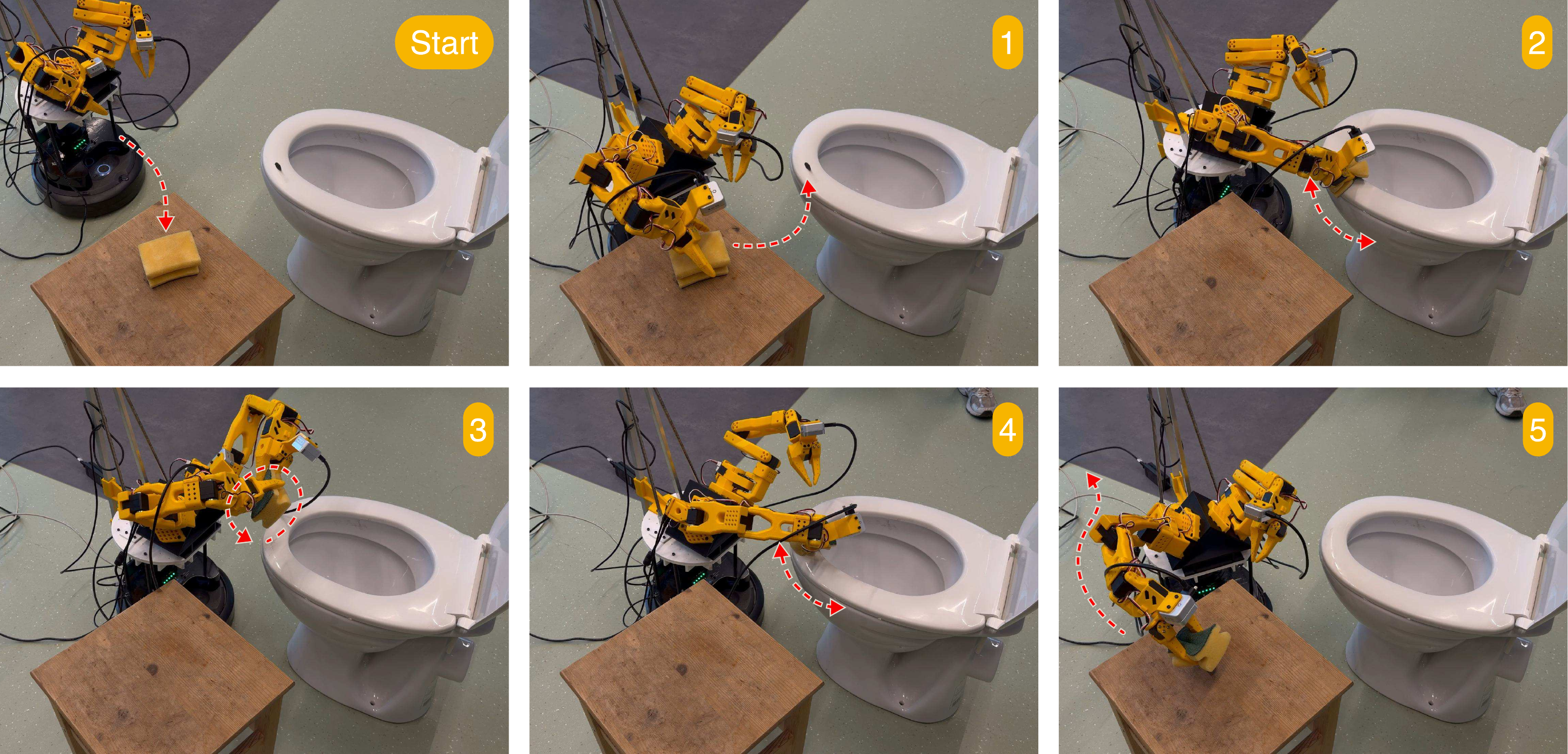}
	\caption{Successful rollout of the toilet-cleaning task, showing the initial state and all five subtasks in temporal order.}
    \label{FigFrames}
\end{figure*}
To assess the practicality of the proposed low-cost mobile bimanual robot, we evaluate four VLA models on a multi-stage toilet-cleaning task. We examine both platform performance and the models' ability to coordinate two arms and a mobile base over long horizons. We describe the task setup and trial randomization (Sec.~\ref{ssec:task_setup}), data collection via teleoperation and human-in-the-loop corrections (Sec.~\ref{ssec:data_collection}), the evaluated policies and two variants for improving subtask transitions and long-horizon awareness (Sec.~\ref{ssec:policy_configurations}), and finally the resulting success rates and failure modes (Sec.~\ref{ssec:results}).

\subsection{Toilet-Cleaning Task and Environment}\label{ssec:task_setup} 
We chose to evaluate the VLA models on a multi-stage bimanual manipulation task, in which the robot removes a dirt smudge from a toilet seat using both sides of a sponge obtained from an adjacent table.
This sanitation task is especially relevant, since it is both of high societal interest for automation~\cite{jiang2024development} and utilizes the capabilities of our robot platform to a high extent.
The task is decomposed into five consecutive subtasks, which can be seen in Fig.~\ref{FigFrames}:
(1)~Grab the sponge and drive to the dirty spot on the toilet seat;
(2)~Clean the dirty spot with the rough sponge side;
(3)~Flip the sponge using both arms;
(4)~Wipe the toilet seat clean with the soft sponge side;
(5)~Put the sponge away and return to the initial position.
This decomposition ensures that the evaluated models fully utilize both arms and the mobile base to clean the toilet.
To this end, the sponge flip performed in subtask 3 is especially significant, since it is the only subtask that strictly cannot be fulfilled unimanually. 
Additionally, this sequential structure allows for a closer comparison of the resulting trajectories and aids in the identification of failure modes.
\par
As shown in Fig.~\ref{FigFrames}, the environment in which the task takes place contains an open toilet with a black stain on the seat, along with a small table holding the sponge.
Both are placed 0.5–1.5 meters in front of the robot's initial position and within the field of view of its cameras.
From the robot's initial position, the table is located to the right of the toilet.
Black food coloring is applied onto the toilet seat to form one line-shaped smudge with a length of 3–6 centimeters and a width of 1–2 centimeters.
In order to ensure the robot can reach the smudge, it is always positioned within the front-facing two-thirds of the seat.
The sponge has a soft and a coarse side, is soaked in soapy water, and is placed on the table with its coarse side facing down.
\par
To introduce environmental variation while keeping the number of required trials manageable, we alter a limited set of task parameters across episodes.
Between trials, we randomize the orientations of the toilet ($\pm45^\circ$) and table (unconstrained), the sponge's position on the table, and the size and placement of the smudge within the front-facing region described above.
On the other hand, parameters such as lighting conditions, toilet model, and background object arrangement are deliberately kept constant.
Similarly, the type and size of the sponge as well as the composition of the soapy water are kept the same.
While greatly simplified in contrast to full real-world conditions, this introduces sufficient variation to explore the generalization capabilities of the policies without requiring substantial environmental changes that would impede data collection.


\subsection{Data Collection}\label{ssec:data_collection}
For training, we collected 250 teleoperation demonstrations at 30 Hz using the LeRobot recording scripts and stored them in the LeRobot dataset format. Each frame contains images from the wrist and head cameras, along with the robot state, comprising the joint states of both arms and the base’s linear and angular velocities. Target actions are derived from the leader arm’s joint configuration. We use absolute actions, although the framework also supports relative actions. All demonstrations share the same high-level task description, while individual frames are manually annotated with subtask labels.

For each demonstration, we randomize the toilet, table, and sponge parameters as described in Sec.~\ref{ssec:task_setup}. Of the 250 demonstrations, 200 are standard teleoperation trajectories (approx. 1.5~min each) and 50 are correction trajectories (approx. 2~min each) collected with LeRobot’s DAgger-style human-in-the-loop framework~\cite{ross2011reduction}. For DAgger collection, we use $\pi_0$ fine-tuned on the initial 200 demonstrations. The operator can seamlessly take over through the teleoperation interface, providing corrective data in out-of-distribution states for subsequent training. Such intervention-based data collection is highly relevant in robotics pipelines and is available out of the box through LeRobot integration.

\subsection{Policy Configurations}\label{ssec:policy_configurations}
We evaluate four vision-language-action policy families on the toilet-cleaning task: $\pi_0$~\cite{black2025pi0}, $\pi_{0.5}$~\cite{black2025pi05}, SmolVLA~\cite{shukor2025smolvla}, and GR00T N1.7~\cite{bjorck2025gr00tn1,nvidia2026gr00tn17}. For all models, we use the LeRobot implementations and fine-tune them on our dataset of 250 demonstrations. In addition to these baselines, we investigate whether providing temporal context improves long-horizon performance, either implicitly through observation memory or explicitly through predicted task-stage information.

It is important to note that $\pi_{0.5}$ is currently only partially open-sourced~\cite{physicalintelligence2024openpi}. The available implementation supports training and inference of the flow-matching action head, while several components described in the original work, including high-level subtask prediction, have not been released. Our evaluation therefore reflects the publicly available implementation rather than the complete $\pi_{0.5}$ system described in the original work.

By default, both $\pi_0$ and $\pi_{0.5}$ condition only on the most recent observation and therefore receive no observation history from which to infer task progression. To provide implicit temporal context, we evaluate $\pi_{0.5}$ with the short-horizon memory mechanism of MEM~\cite{torne2026mem} implemented in LeRobot. We use six memory frames with a stride of 30, providing snapshots spanning the preceding six seconds. Temporal attention is applied every four layers, and proprioceptive information is included in memory. This configuration tests whether recent observation history helps the policy infer task progression and determine when to transition between subtasks.

To provide explicit temporal context, we evaluate a stage-aware variant of $\pi_0$ using signals derived from SARM~\cite{chen2026sarm}. As described earlier, we augment the observation with five subtask scores in $[0,1]$, representing the estimated likelihood of the robot being in each subtask, together with an overall task-completion estimate in $[0,1]$.

To generate these signals for policy training, we fine-tune SARM on the manually labeled subtasks in our dataset using five-fold cross-validation. For each fold, SARM is trained on four folds and used to predict the subtask and progress signals for the held-out fold. We then combine the out-of-fold predictions from all five folds to construct the augmented training dataset for $\pi_0$. This ensures that the stage signals used for policy training are generated for samples that were not seen during the corresponding SARM training, avoiding unrealistically accurate in-sample predictions. For policy evaluation, we train a final SARM model on the complete training set and use its predictions online as input to $\pi_0$.

Together, the two variants test complementary approaches to long-horizon task awareness: inferring task progression implicitly from recent observation history and providing it explicitly through predicted task-stage information.

\subsection{Results}\label{ssec:results}
\begin{figure}[htbp]
    \centering
	\includegraphics[width=0.8\columnwidth]{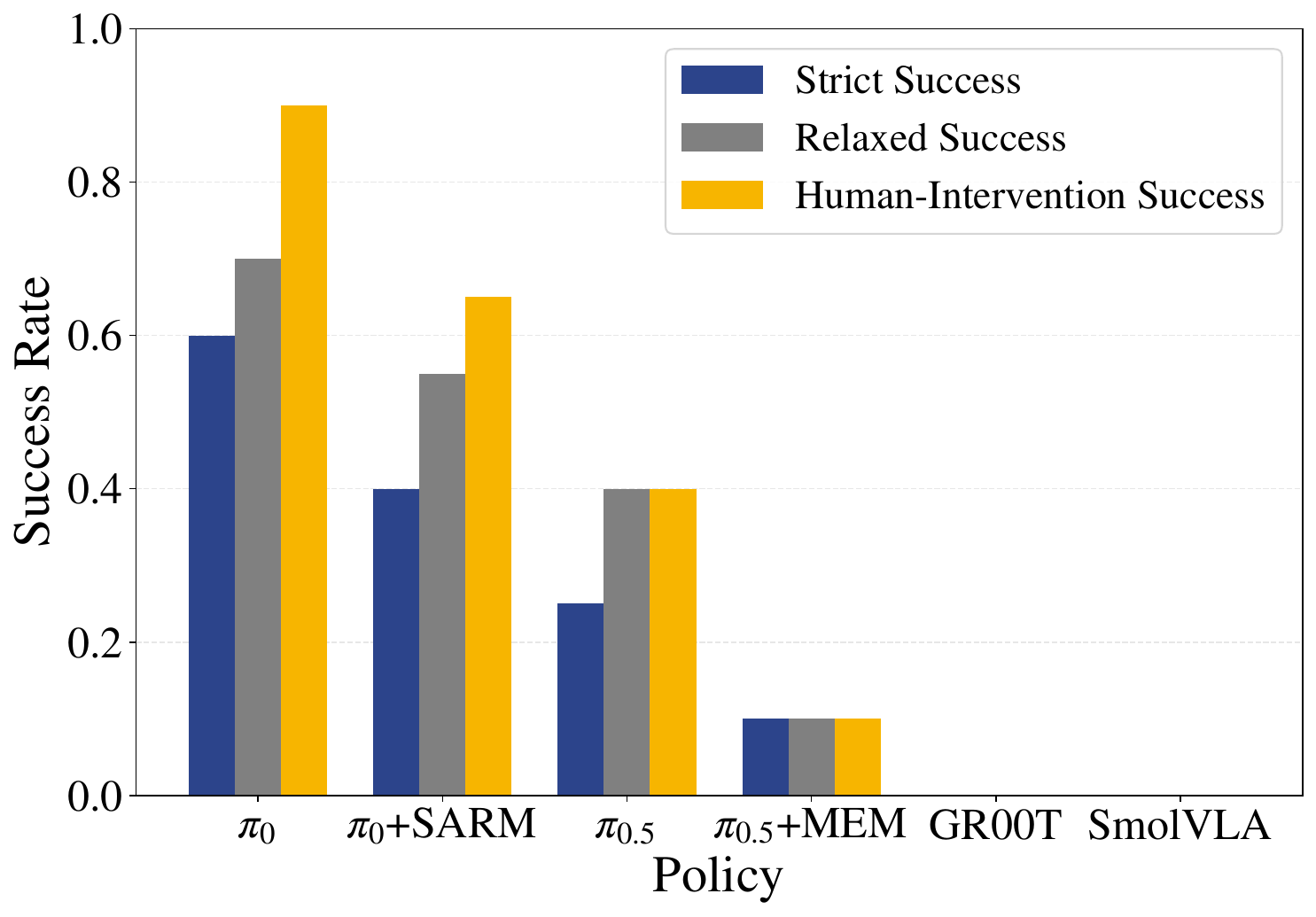}
	\caption{Success rates of the different policies for the toilet cleaning task.}
    \label{FigPolicySuccessRates}
\end{figure}

For evaluation, we distinguish three success criteria. Strict success requires complete removal of the dirty spot, whereas relaxed success allows small residual spots that are barely visible in the camera images and are therefore attributed primarily to hardware limitations rather than policy failure. We additionally report human-intervention success. Several policies occasionally entered cyclic or deadlocked states and reached the per-stage timeout of three minutes. In these cases, a human repositioned the robot into an in-distribution state without otherwise controlling the task. At most one such intervention was allowed per rollout; if the robot became stuck again, the rollout was counted as a failure. If the policy subsequently completed the task autonomously, the rollout was counted as a human-intervention success. We evaluate 20 rollouts per policy, except for GR00T, for which all five evaluated rollouts failed due to fundamental policy issues that prevented successful task execution. Figure~\ref{FigPolicySuccessRates} summarizes the quantitative results.

Among the evaluated policies, $\pi_0$ performs best across all success criteria, achieving a strict success rate of 60\%. Its main failure modes are getting stuck under the toilet seat, or failing to transition to the next subtask. We did not observe failures in sponge grasping or retention. This results in a relaxed success rate of 70\%, which increases to 90\% with a single human intervention to recover from stuck states.

We evaluated $\pi_0$ with SARM to determine whether providing information about overall task and subtask progress could reduce failures from cycling behavior or failing to transition between phases. This was not the case: $\pi_0$ with SARM performed worse than the baseline, achieving strict, relaxed, and human-intervention success rates of 40\%, 55\%, and 65\%, respectively.

The successor model, $\pi_{0.5}$, performed worse than $\pi_0$ in our experiments. It frequently became stuck near the toilet or missed the target spot. Additionally, its sponge grasping was less reliable, occasionally causing the sponge to be dropped during handover. It achieved strict and relaxed success rates of 25\% and 55\%, respectively. Notably, human intervention did not lead to additional successful completions in any rollout.

To investigate whether memory could alleviate $\pi_{0.5}$'s tendency to get stuck, we evaluate LeRobot's implementation of $\pi_{0.5}$ with MEM~\cite{torne2026mem}. LeRobot currently implements only the short-horizon memory component of MEM. Since pretrained weights for this component were not released, it is fine-tuned solely on our dataset. In our experiments, this did not improve performance: all success criteria reached only 10\%. The policy continued to exhibit the previously observed failure modes, including getting stuck, dropping the sponge, and failing to transition between stages.

SmolVLA is particularly interesting because it is primarily trained on community data collected with SO-100 and SO-101 arms. In our experiments, the policy struggled to localize the dirt on the toilet. Unlike the previous models, which generally moved into the vicinity of the target spot, SmolVLA often failed to approach it and in most rollouts never reached the sponge. It achieved a 0\% success rate across all categories. Human intervention was rarely applicable because the policy typically did not get stuck, but instead completed its rollout without cleaning the toilet.

Finally, GR00T struggled with the basic requirement of picking up the sponge and showed no indication of being able to complete the task. It therefore achieved a 0\% success rate across all categories.




\section{Discussion}
We discuss the case study results and their implications for robot design.

A key finding is that $\pi_0$ outperforms $\pi_{0.5}$ on the mobile toilet-cleaning task. One possible explanation is that the released $\pi_{0.5}$ implementation differs from the full model described in the original work, as discussed in Sec.~\ref{ssec:policy_configurations}, and only includes the flow-matching head~\cite{physicalintelligence2024openpi}. Consequently, the high-level/low-level separation of the full model is absent. This architectural difference may contribute to the observed performance gap, although our experiments do not isolate its effect. Both policies became stuck in similar situations, but human intervention allowed $\pi_0$ to recover, whereas $\pi_{0.5}$ typically returned to the same stuck state, suggesting differences in their ability to recover from out-of-distribution states.

Both adapted models, $\pi_0$ with SARM and $\pi_{0.5}$ with memory, perform worse than their respective baselines. One possible explanation is that introducing additional state information with previously unseen values during fine-tuning complicates adaptation. Determining whether the degradation originates from the additional conditioning, the fine-tuning procedure, or other factors would require further ablations.

SmolVLA's limited performance may be related to the lack of mobile bimanual manipulation data in its training distribution. This highlights a broader data gap that our platform could help address by enabling the collection of community datasets for mobile bimanual manipulation. Expanding such datasets could make lightweight policies such as SmolVLA more suitable for compute-constrained mobile manipulators.

GR00T, in contrast, struggled to control the new embodiment and failed at basic parts of the task, such as grasping the sponge. Its focus on humanoid embodiments may contribute to this difficulty, suggesting that embodiment transfer remains a significant challenge when applying generalist policies to substantially different robot platforms.

The proposed robot provides an out-of-the-box pipeline for data collection, policy training, and policy deployment. Wider community adoption could further enable shared datasets, similar to those that emerged around the SO-101 platform. This lowers the barrier to prototyping by providing an affordable platform for experimentation before committing substantial time and financial resources.

The platform is also well suited for higher education. Mobile manipulators commonly used for physical intelligence research can cost more than \texteuro{}25\,000 per unit~\cite{gros2023unlocking}, making courses with multiple robots impractical. In contrast, our system costs approximately \texteuro{}3200, depending on the camera setup and TurtleBot configuration, allowing multiple units to fit within a typical university teaching budget. For institutions that already own TurtleBot~4 platforms, the barrier is even lower, as the manipulator extension itself costs only approximately \texteuro{}500. Combined with the integrated software pipeline, this allows students to focus on robotics and learning problems rather than on setting up the underlying infrastructure.

We selected the SO-101 arms because they are widely adopted as low-cost manipulators. However, their 5-DOF design and limited reach constrain the range of tasks the platform can perform, particularly those requiring a larger workspace or greater dexterity. These limitations can be addressed through future design iterations. Importantly, the modular design allows the SO-101 arms to be replaced with alternative manipulators of higher capabilities, without redesigning the entire platform.


\section{Conclusion}

In this work, we investigated whether an accessible, low-cost mobile
bimanual robot can provide a practical platform for long-horizon
cleaning and VLA research. We developed a mobile bimanual system
combining two LeRobot SO-101 manipulators with a TurtleBot~4, custom
3D-printable structures, multi-camera perception, and a modular
LeRobot--ROS~2 software stack. The platform supports the complete
workflow from teleoperation and data collection to policy training,
human-in-the-loop correction, and physical deployment.

As a proof of concept, we studied toilet cleaning, a representative
household task that combines mobility, bimanual coordination,
contact-rich manipulation, tool use, and long-horizon execution.
Using 250 real-world demonstrations, we evaluated several
representative publicly available VLA implementations under the same embodiment, dataset,
and physical conditions. Among them, $\pi_0$ showed the strongest
performance after adaptation to our mobile bimanual setting, achieving 60\% strict
and 70\% relaxed autonomous success. With limited human intervention
to recover from dead ends, 90\% of its rollouts could complete the
task.

Our experiments also illustrate that long-horizon reliability cannot
be obtained simply by providing additional temporal information.
Neither SARM-derived task-progress signals nor short-term MEM
observation history improved their corresponding VLA baselines in our
setting. Across the evaluated policies, failures in embodiment
adaptation, task transitions, and recovery from out-of-distribution
states remain important obstacles to reliable autonomous execution.

These results demonstrate the feasibility of studying complex mobile
bimanual cleaning behaviors using accessible hardware, while
highlighting the remaining limitations of current open VLA policies.
Rather than replacing higher-performance robotic systems, the
presented platform provides a reproducible testbed on which
demonstrations can be collected, policies can be rapidly evaluated,
and failure modes can be studied before transferring promising
approaches to more capable real-world platforms. Its low-cost,
open-source design may additionally facilitate community-scale
collection of mobile bimanual data and make long-horizon robot
learning more accessible for research and education.

\section{Limitations and Future Work}

While the presented platform demonstrates the feasibility of
long-horizon mobile bimanual manipulation on accessible hardware,
several limitations remain. First, our physical evaluation focuses on
a single toilet-cleaning scenario. Although object poses and the
location and size of the dirt are randomized, other environmental
factors, including lighting, background, toilet model, and cleaning
tools, remain fixed. Evaluating additional household tasks and more
diverse environments will therefore be necessary to better
characterize the generalization capabilities of both the platform and
the evaluated VLA policies.

Second, the current hardware represents a proof-of-concept design and
can be further improved. The mechanical structure is primarily
designed around lightweight SO-101 manipulators and 3D-printed
components. Future versions could adopt more capable open-source
6-DoF manipulators and reinforce the supporting structure to increase
payload capacity and mechanical stability. The current system also
relies on external power and workstation-based policy inference.
Integrating a portable power supply and more capable onboard computing
would enable a more self-contained platform and facilitate deployment
over larger environments.

Finally, the current study is limited to physical experiments and a
relatively small real-world dataset. A corresponding simulation
environment could support scalable policy development, systematic
evaluation, and sim-to-real studies before deployment on the physical
robot. Combined with evaluation across a broader range of mobile
bimanual tasks, this would also enable larger-scale collection of
mobile bimanual manipulation data and provide a more comprehensive
benchmark for studying VLA generalization and long-horizon execution.



\bibliographystyle{IEEEtran}
\bibliography{ref}

\end{document}